\documentclass[twoside]{article}

\usepackage{amsmath}
\usepackage{hyperref}
\usepackage{geometry}
\usepackage{makecell}
\usepackage{xcolor}
\usepackage{amsfonts}
\usepackage{booktabs}
\usepackage{amssymb}   
\usepackage{subcaption} 
\usepackage[ruled,vlined]{algorithm2e}
\usepackage{todonotes}
\usepackage{wrapfig} 
\usepackage{microtype} 

\usepackage{titlesec}
\titleformat{\section}
  {\normalfont\Large\bfseries}  
  {\thesection}                 
  {1em}                         
  {\MakeUppercase}               

\usepackage[round]{natbib}

\begin{document}

\title{DelTriC: A Novel Clustering Method with Accurate Outlier Detection}
 \author{Tomáš Javůrek, Michal Gregor, Sebastian Kula, Marian Simko \\
 Kempelen Institute of Intelligent Technologies}
\maketitle





\begin{abstract}
The paper introduces \textbf{DelTriC} (\textbf{Del}aunay \textbf{Tri}angulation \textbf{C}lustering), a clustering algorithm which integrates PCA/UMAP-based projection, Delaunay triangulation, and a novel back-projection mechanism to form clusters in the original high-dimensional space. DelTriC decouples neighborhood construction from decision-making by first triangulating in a low-dimensional proxy to index local adjacency, and then back-projecting to the original space to perform robust edge pruning, merging, and anomaly detection. DelTriC can outperform traditional methods such as k-means, DBSCAN, and HDBSCAN in many scenarios; it is both scalable and accurate, and it also significantly improves outlier detection.



\end{abstract}

\section{Introduction}

State-of-the-art clustering algorithms suffer from either poor scalability or sensitivity to data shape and dimensionality. For instance, k-means~\cite{kmeans} performs well only on convex, blob-like clusters and requires the number of clusters to be predefined. DBSCAN, while capable of detecting arbitrarily shaped clusters, is sensitive to parameter selection and struggles in high-dimensional spaces~\cite{dbscan}.
 
HDBSCAN~\cite{hdbscan}, while more robust to noise and cluster shape, suffers from significant computational overhead in high-dimensional settings and when processing large datasets. Practical speed-up strategies exist, such as applying dimensionality reduction (e.g., PCA~\cite{pca}, UMAP~\cite{umap}) prior to clustering, using approximate nearest neighbor search (e.g., \cite{dong2011efficient}: the method implemented in \href{https://github.com/lmcinnes/pynndescent}{PyNNDescent}, FAISS~\cite{faiss}), or clustering a representative sample of the data followed by classification of the remaining points. However, these approaches may lose structural information when the separability of clusters arises from patterns that only manifest in the full high-dimensional representation, rather than in the low-dimensional embedding~\cite{Heckel2014, Rosito2023}. This trade-off between speed and fidelity remains a fundamental challenge motivating the search for more efficient and accurate clustering algorithms.

To address these limitations, we propose DelTriC, which integrates PCA/UMAP-based projection, Delaunay triangulation, and a novel back-projection mechanism to form clusters in the original space.
This approach decouples neighborhood construction from decision-making: it first triangulates in a low-dimensional proxy to index local adjacency, then back-projects to the original space to perform robust edge pruning, merging, and anomaly detection where separation signals actually reside.

The detailed procedure is described in Section~\ref{sec:algorithm}, and an illustrative walkthrough is provided in Appendix~\ref{app:deltric}.


\section{Related Work}
\label{relatedwork}

Triangulation-based clustering methods have become increasingly popular for handling spatial data with irregular shapes, heterogeneous densities, and complex domains. Here, we compare the proposed DelTriC algorithm with several recent triangulation-based clustering approaches.

\textbf{nDTAg (Debbad et al., 2025)~\cite{debbad2025ndtag}:}
nDTAg uses Delaunay triangulation to classify edges as “long” or “short” and forms clusters by grouping simplices with only short edges. It then merges clusters based on directional variance and reassigns unclassified points. While nDTAg is effective for non-uniform densities and non-linear partitions, it operates directly in the triangulated space and does not explicitly address the separation of adjacent clusters with heterogeneous densities. Importantly, nDTAg does not implement a back-projection mechanism to recover clusters in the original multidimensional space; its clustering is performed within the triangulation graph, which may limit interpretability and flexibility in high-dimensional settings.

\textbf{Delaunay Graph Descending Clustering (Qiu \& Li, 2015)~\cite{qiu2015delaunay}:}
Qiu and Li propose organizing data into in-tree structures within the Delaunay graph, using a physically-inspired “descending to the nearest neighbor” rule. Their method avoids redundant edges and automates cluster formation, but it is sensitive to parameter choices and does not incorporate density-based partitioning or explicit noise handling as in DelTriC. Like nDTAg, this method operates within the triangulation graph and does not provide a systematic back-projection to the original data space, which can be problematic for multidimensional data analysis.

\textbf{Adaptive DPC-DG (Xingqiong \& Kang, 2025)~\cite{xingqiong2025adaptive}:}
DPC-DG combines Delaunay graphs with adaptive density peak clustering, automatically determining cluster centers and cutoff distances. While DPC-DG is robust to arbitrary cluster shapes and density variations, it does not use triangulation for density estimation or for the explicit removal of global/local effects, which are central to DelTriC’s improved performance on adjacent and nested clusters. DPC-DG can be applied in multidimensional settings, but it does not feature a dedicated back-projection step; clustering is performed in the graph space, and the mapping back to the original data is implicit rather than explicit.

\textbf{DTSCAN (K. Kim and J. Cho.)~\cite{dtscan}:}
DTSCAN (Delaunay Triangulation-based SCAN) is a clustering algorithm that enhances the classical DBSCAN by leveraging the connectivity structure of Delaunay triangulation. Instead of using a fixed-radius neighborhood, DTSCAN defines neighborhoods based on the adjacency of points in the triangulation graph, allowing the method to adapt to local data geometry and better capture clusters with irregular shapes or varying densities. This approach improves the separation of adjacent clusters and addresses the “touching” problem that often challenges traditional density-based methods. However, DTSCAN does not incorporate advanced density estimation, density-based partitioning, or explicit back-projection to the original multidimensional space – key innovations introduced by DelTriC. As a result, while DTSCAN improves upon standard DBSCAN in spatial settings, it may still struggle with highly heterogeneous densities or complex, high-dimensional domains, where DelTriC excels~\cite{zhou2025triangulation}.

All these efforts demonstrate that Delaunay triangulation is a powerful and versatile tool for clustering, with the potential to become a cornerstone technique in future methods. However, none of the existing studies explicitly address the quality of anomaly detection alongside clustering. Since our work is motivated by the dual objective of preserving strong clustering performance while simultaneously enabling principled outlier detection, a direct comparison would be misleading.

\section{The Algorithm}
\label{sec:algorithm}

DelTriC is a robust statistical edge-pruning, and centroid-based merging algorithm to detect clusters in high-dimensional datasets. The algorithm is designed to preserve local geometry while robustly eliminating spurious connections and handling anomalies. This section provides a mathematical description of the DelTriC algorithm, with particular emphasis on its principal innovation: the back-projection step, which enables robust clustering in high-dimensional spaces. Additionally, we introduce a median-based normalization (MAD) within the edge pruning mechanism – a detail that, to our knowledge, has not previously been applied in this context. The intuition and motivation behind these steps are further illustrated in Appendix~\ref{app:deltric}. For further details on the mathematical motivation of DelTriC, see Appendix~\ref{theory}.

\subsection{Dimensionality Reduction and Triangulation}

Let $X \in \mathbb{R}^{n \times d}$ denote the dataset with $n$ points in $d$ dimensions.  
To enable efficient triangulation, we project $X$ to a two-dimensional embedding $X_{\text{proj}} \in \mathbb{R}^{n \times 2}$ using either \textit{UMAP} or \textit{PCA}:
\[
X_{\text{proj}} =
\begin{cases}
\text{PCA}_2(X), & \text{if method = PCA}, \\
\text{UMAP}_2(X), & \text{if method = UMAP}.
\end{cases}
\]

On $X_{\text{proj}}$, we compute the \textit{Delaunay triangulation} $\mathcal{T}$, which yields a set of triangles $\Delta = \{t_1, \dots, t_m\}$ with vertex indices $(i,j,k)$.  
For each triangle $t$, we compute the edge lengths in both the original space $X$ and the projection $X_{\text{proj}}$:
\[
\ell_{ab} = \| X_a - X_b \|_2, \quad 
\ell^{\text{proj}}_{ab} = \| X^{\text{proj}}_a - X^{\text{proj}}_b \|_2.
\]
The \textit{triangle size} is by default defined as the maximum edge length:
\begin{align*}
s(t) &= \max \{ \ell_{ij}, \ell_{jk}, \ell_{ki} \}, \\
s_{\text{proj}}(t) &= \max \{ \ell^{\text{proj}}_{ij}, \ell^{\text{proj}}_{jk}, \ell^{\text{proj}}_{ki} \}.
\end{align*}
but sum or a shortest mutual distance can be used as well.

\subsection{Sigma-based Edge Pruning}

To distinguish dense regions from noise, we normalize triangle sizes using the median and the Median Absolute Deviation (MAD):
\begin{align*}
z(t) &= \frac{s(t) - \mathrm{median}(s)}{\mathrm{MAD}(s)}, \\
z_{\text{proj}}(t) &= \frac{s_{\text{proj}}(t) - \mathrm{median}(s_{\text{proj}})}{\mathrm{MAD}(s_{\text{proj}})}.
\end{align*}
A pruning threshold $\theta$ is defined as
\[
\theta = \mu + \sigma_f \cdot \sigma,
\]
where $\mu$ and $\sigma$ are the mean and standard deviation of $z(t)$, and $\sigma_f$ is a user-defined pruning factor. The use of \textbf{median}-based normalization combined with a \textbf{mean}-based threshold is intentional: 
the median and MAD are robust to skewness and outliers, making them suitable for re-scaling triangle sizes in cases where their distribution is asymmetric. 
The subsequent mean and standard deviation provide a scale-sensitive measure that adapts to the global distribution of normalized sizes. 
This hybrid approach ensures that both local robustness and global sensitivity are maintained.  

A triangle $t$ is retained if
\[
\max(z(t), z_{\text{proj}}(t)) \leq \theta,
\]
otherwise it is discarded. The edges of discarded triangles are removed, and the remaining edges form a graph $G=(V,E)$. 
The connected components of $G$ define the set of initial clusters.

This pruning strategy has a dual effect. 
On the one hand, it preserves the manifold structure revealed by the UMAP projection, ensuring that intrinsic geometry of the data is respected. 
On the other hand, by explicitly reacting to oversized triangles that may appear even inside a projected manifold, it exposes anomalous points and spurious connections that would otherwise remain hidden. 
This interplay between manifold preservation and anomaly sensitivity is a key innovation of our algorithm and, as demonstrated in this paper, leads to strong anomaly detection capabilities.

The remaining edges form a graph $G=(V,E)$ with vertices $V=\{1,\dots,n\}$. The connected components of $G$ yield the set of initial clusters.

\subsection{Cluster Merging via Centroid Triangulation}

Let $\mathcal{C} = \{C_1, \dots, C_k\}$ denote the initial clusters. For each cluster $C_i$, we select a representative point $r_i$ as the member closest to the projected mean:
\[
r_i = \arg \min_{p \in C_i} \| X^{\text{proj}}_p - \frac{1}{|C_i|} \sum_{q \in C_i} X^{\text{proj}}_q \|_2.
\]
The set of representatives $\{r_1, \dots, r_k\}$ is triangulated, and sigma-pruning is applied again.  
Clusters connected by surviving edges are merged, yielding larger, coherent groups. In practice, this merging step tends to be \emph{conservative}. 
When optimal hyperparameters are used, it merges only a small number of closely related subclusters, while leaving the majority of clusters unchanged. 
Thus, the main structural partitioning is governed by the initial sigma-pruning stage, and the centroid-based merging acts primarily as a refinement step that corrects over-fragmentation rather than aggressively combining clusters.

\subsection{Anomaly Detection and Merging}
Clusters with size $|C_i| \leq \tau$ (typically $\tau=2$ or $3$) are considered anomalies denoted $A$.  
For each anomaly point $p$ and neighbor $q$ we compute distances using both the original and projected spaces:
\[
d_{pq} = \max\left( \alpha \|X_p - X_q\|_2,\; \|X^{\text{proj}}_p - X^{\text{proj}}_q\|_2 \right),
\]
where $\alpha = \frac{\mathrm{median}(s_{\text{proj}})}{\mathrm{median}(s)}$ is a scaling factor that balances distance scales.  

A cluster-level score is then defined as
\[
S(C_j \mid A) = \frac{\mu}{n} \sum_{p \in A} \sum_{q \in C_j} \frac{1}{d_{pq}}.
\]

where $n$ is the total number of direct neighbors of all points $p\in A$ and $\mu$ is the average triangle size acting as a scale parameter.
The anomaly is merged into the cluster $C_j$ with maximal score, provided $S(C_j \mid A) > \delta$, where $\delta$ is an anomaly sensitivity threshold. In most of the experiments reported in this paper, the value of $\delta$ is fixed in order to maximize the number of anomalies retained, thereby emphasizing the algorithm’s anomaly detection capability.  
If no suitable cluster is found under this criterion, $A$ remains labeled as anomaly.

\subsection{Final Label Assignment}

Each point inherits the label of its cluster. Anomalies that cannot be merged are assigned label $-1$.

\subsection{Pseudocode}

A pseudocode of the DelTriC pipeline is provided in the Algorithm~\ref{alg}.


\begin{algorithm}[tb]
\caption{DelTriC: Triangulation-based Sigma Clustering}
\label{alg}
\KwIn{
    Dataset $X \in \mathbb{R}^{n \times d}$\;
    Pruning threshold $\sigma_f$\;
    Merging threshold $\lambda$\;
    Anomaly size threshold $\tau$
}
\KwOut{Cluster labels $y \in \{-1, 0, 1, \dots\}$}

\SetKwFunction{Proj}{Project}
\SetKwFunction{Triangulate}{DelaunayTriangulation}
\SetKwFunction{Prune}{PruneEdges}
\SetKwFunction{Components}{ConnectedComponents}
\SetKwFunction{Represent}{ComputeRepresentatives}
\SetKwFunction{Merge}{MergeClusters}
\SetKwFunction{Anomaly}{DetectAnomalies}
\SetKwFunction{Label}{AssignLabels}

$X_{\text{proj}} \gets$ \Proj{$X$}\tcp*{using UMAP/PCA}
$\mathcal{T} \gets$ \Triangulate{$X_{\text{proj}}$}\;
$G \gets$ \Prune{$\mathcal{T}, \sigma_f$}\;
$\mathcal{C} \gets$ \Components{$G$}\;
$\{r_i\} \gets$ \Represent{$\mathcal{C}$}\;
$\mathcal{C} \gets$ \Merge{$\mathcal{C}, \lambda$}\;
$\mathcal{C} \gets$ \Anomaly{$\mathcal{C}, \tau$}\;
$y \gets$ \Label{$\mathcal{C}$}\;
\Return $y$\;
\end{algorithm}

\section{Time Complexity}
The time complexity of DelTriC in its default form is approximately $\mathcal{O}(d^2 n) + \mathcal{O}(n \log n)$, where $d$ is the number of dimensions and $n$ is the number of data points. This is due to:
\begin{itemize}
    \item Dimensionality Reduction:
    \begin{itemize}
        \item PCA: $\mathcal{O}(d^2 n)$ (worst case scenario, not optimized)
        \item UMAP: $\mathcal{O}(d n^2)$ (worst case scenario, not optimized)
    \end{itemize}
    \item Delaunay triangulation in 2D: $\mathcal{O}(n \log n)$,
    \item Back-projection and triangle filtering: $\mathcal{O}(d \cdot n)$.
\end{itemize}

The comparison of time consumption by individual algorithms used in this paper is briefly discussed in Appendix \ref{time_complexity}.

\section{Benchmarking and Evaluation}

\subsection{Data Generation}

Synthetic datasets were generated with a custom function based on \texttt{sklearn.datasets.make\_blobs}. 
The function allows us to control the number of points (\texttt{n\_points}), number of clusters (\texttt{n\_clusters}), 
dimensionality (\texttt{n\_dim}), expected cluster overlap (\texttt{overlap}), and the fraction of anomalies 
(\texttt{anomaly\_fraction}). Randomness was controlled with a fixed \texttt{random\_state}. 

The standard deviation of the Gaussian blobs was scaled with dimensionality to control overlap:
\[
\texttt{cluster\_std} = \texttt{overlap} \times \sqrt{n_{\text{dim}}} \times 2.0 .
\]

By default, approximately 20\% of the points are assigned to an additional irregular ``snake'' cluster. 
This cluster was generated by sampling from a sinusoidal curve with Gaussian noise,
\[
x = \sin(t) + \varepsilon, \quad y = \cos(t) + \varepsilon,
\]
for $t \in [0,4\pi]$, scaled and shifted to match the distribution of the blobs. 
For higher dimensions, additional coordinates were filled with small Gaussian noise. 
This irregular cluster receives its own label distinct from the Gaussian blobs.

Finally, anomalies were added as uniformly distributed points within the bounding box of the dataset. 
These anomalies were labeled with $-1$. 

The function can be summarized as: Gaussian blobs $+$ one irregular snake-shaped cluster $+$ uniformly scattered anomalies.
This design provides a balance between standard clustering scenarios and more challenging irregular structures.

\subsection{Scoring}

We report two standard clustering evaluation metrics:

\begin{itemize}
    \item \textbf{Adjusted Rand Index (ARI)}: Measures the similarity between the predicted and true cluster assignments, adjusted for chance. It ranges from -1 to 1, with 1 indicating perfect agreement.
    \item \textbf{Normalized Mutual Information (NMI)}: Captures the mutual dependence between predicted and true labels, normalized to range from 0 (no mutual information) to 1 (perfect correlation).
\end{itemize}

In the case of the outlier detection performance evaluation we use $f1$-score targeting to the anomaly class (-1).

\subsection{Hyperparameter Search Space}

Instead of manually defining discrete grids of hyperparameters, we employ
\textbf{Optuna}, which leverages the Tree-structured Parzen Estimator (TPE) 
to perform efficient hyperparameter optimization. Optuna dynamically explores 
continuous intervals and categorical choices, allowing for a more flexible 
search strategy compared to fixed grids.

The following search spaces were used for each algorithm:

\begin{table}[h]
\centering
\begin{tabular}{|l|l|}
\hline
\textbf{Algorithm} & \textbf{Search Space} \\
\hline
HDBSCAN & 
$\epsilon \sim \mathcal{U}(0.0, 0.6)$, \\
& selection\_method $\in \{\text{eom}, \text{leaf}\}$, \\
& $n\_components = 10$, \\
& dim\_reduction = \texttt{umap} \\
\hline
DBSCAN & 
$eps \sim \mathcal{U}(0.001, 1.0)$, \\
& $min\_samples \sim \mathcal{U}\{3, 20\}$, \\
& $n\_components = 10$, \\
& dim\_reduction = \texttt{umap} \\
\hline
DelTriC & 
$prune\_param \sim \mathcal{U}(0.5, 2.0)$, \\
& $merge\_param \sim \mathcal{U}(-2.0, -0.5)$, \\
& dim\_reduction = \texttt{umap} \\
\hline
KMeans & 
$n\_clusters = n$ (true \# of clusters) \\
\hline
Spectral & 
$n\_clusters = n$ (true \# of clusters) \\
\hline
\end{tabular}
\caption{Optuna search space for each algorithm. Continuous ranges are sampled uniformly, categorical parameters are sampled from discrete sets.}
\end{table}

The hyperparameters selected for optimization were carefully chosen for each model to address the issue of limited cluster separation capability, as demonstrated in subsequent sections. Specifically, each model was independently and rigorously optimized to ensure optimal performance (the best ARI score). In practice, UMAP is often used to reduce data to 10–20 dimensions before applying HDBSCAN. This range provides a good trade-off between preserving the intrinsic cluster structure and maintaining computational efficiency, as HDBSCAN’s runtime and memory usage grow with dimensionality, making higher-dimensional embeddings slower and more resource-intensive.

\section{Performance on Synthetic Data}

Clustering accuracy was evaluated using the Adjusted Rand Index (ARI) and Normalized Mutual Information (NMI), as shown in Table~\ref{tab:synthetic_new}. For DBSCAN and HDBSCAN, UMAP with 10 components was applied when the dimensionality of the generated data exceeded 10. DelTriC, by default, employs UMAP with dimensionality reduction to two components. The generated datasets are almost ideal for centroid-based algorithms such as \textit{k}-means, with the exception of one irregular cluster and the presence of anomalies.

Local hyperparameter optimization was performed for all algorithms. However, for \textit{k}-means and spectral clustering, the number of clusters~$k$ was fixed to the true number of generated clusters in order to ensure optimal performance.


\begin{table}[htbp]
\centering
\caption{Comparison of Clustering Algorithms Performance on Synthetic Data (ARI \textbar NMI)}
\small
\label{tab:synthetic_new}
\begin{tabular}{lllll} 
\toprule
$n_{\text{points}}$ & dim & DBSCAN & DelTriC & HDBSCAN \\
\midrule
500  & 5   & \textbf{0.58} \textbar 0.62 & 0.56 \textbar \textbf{0.63} & 0.54 \textbar 0.60 \\
500  & 20  & 0.60 \textbar 0.64 & 0.59 \textbar 0.65 & 0.61 \textbar 0.65 \\
500  & 50  & 0.21 \textbar 0.35 & 0.20 \textbar 0.35 & 0.16 \textbar 0.33 \\
5000 & 5   & 0.57 \textbar 0.62 & 0.47 \textbar 0.56 & \textbf{0.58} \textbar \textbf{0.60} \\
5000 & 20  & \textbf{0.81} \textbar 0.78 & 0.71 \textbar 0.74 & 0.79 \textbar 0.77 \\
5000 & 50  & 0.57 \textbar 0.63 & \textbf{0.73} \textbar 0.77 & 0.61 \textbar 0.66 \\
10000 & 5  & \textbf{0.66} \textbar \textbf{0.64} & 0.52 \textbar 0.58 & 0.54 \textbar 0.58 \\
10000 & 20 & \textbf{0.80} \textbar \textbf{0.78} & 0.73 \textbar 0.76 & 0.72 \textbar 0.73 \\
10000 & 50 & 0.70 \textbar 0.71 & \textbf{0.81} \textbar \textbf{0.82} & 0.70 \textbar 0.72 \\
\midrule
$n_{\text{points}}$ & dim & KMeans & Spectral & \\  
\midrule
500  & 5   & 0.53 \textbar 0.59 & 0.55 \textbar 0.62 & \\
500  & 20  & \textbf{0.68} \textbar \textbf{0.71} & 0.56 \textbar 0.62 & \\
500  & 50  & \textbf{0.85} \textbar \textbf{0.85} & 0.30 \textbar 0.37 & \\
5000 & 5   & 0.51 \textbar 0.55 & 0.10 \textbar 0.33 & \\
5000 & 20  & 0.80 \textbar \textbf{0.79} & 0.45 \textbar 0.62 & \\
5000 & 50  & 0.68 \textbar \textbf{0.79} & 0.57 \textbar 0.60 & \\
10000 & 5  & 0.52 \textbar 0.56 & 0.41 \textbar 0.57 & \\
10000 & 20 & 0.79 \textbar 0.78 & 0.45 \textbar 0.62 & \\
10000 & 50 & 0.69 \textbar 0.79 & 0.59 \textbar 0.63 & \\
\bottomrule
\end{tabular}
\end{table}

\begin{figure}[h]
    \centering
    \includegraphics[width=0.8\columnwidth]{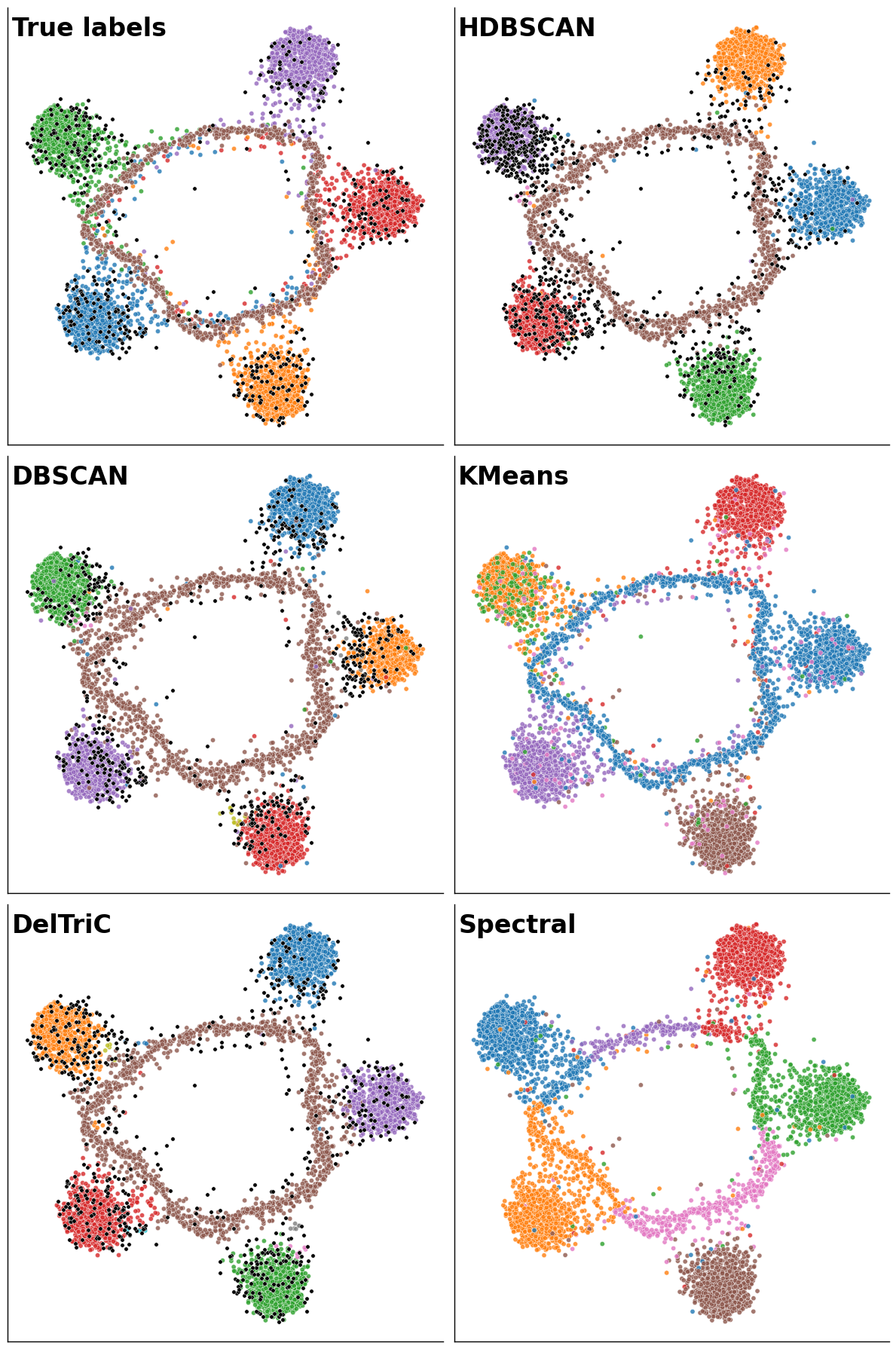}
    \caption{True Labels and all tested algorithms on benchmark $n_{\text{points}}=5000$ and $\text{dim}=50$}
    \label{fig:benchmark}
\end{figure}

The results in Table~\ref{tab:synthetic_new} demonstrate the strong performance of DelTriC, comparable to DBSCAN and HDBSCAN. They further indicate that DelTriC may outperform these two algorithms in higher dimensions, for example in the benchmark with $n_{\text{points}}=5000$ and $\text{dim}=50$, as shown in Figure~\ref{fig:benchmark}. Interestingly, \textit{k}-means achieves stronger results for smaller sample sizes but higher dimensionalities, which may be attributed to its better mitigation of the curse of dimensionality when point densities are lower.

To evaluate the sensitivity of the individual algorithms to anomalies, we report the F1-scores for the same benchmarks in Table~\ref{tab:synthetic_anomaly}. DelTriC clearly dominates this test, highlighting its key strength: \textbf{while maintaining clustering performance comparable to HDBSCAN and DBSCAN, it provides substantially higher accuracy in anomaly detection}. This can also be observed in the benchmark plot in Figure~\ref{fig:benchmark}, where DBSCAN and HDBSCAN struggle to identify anomalies (depicted in dark gray) embedded deeper inside the clusters. In contrast, comparison of the true labels with the DelTriC output shows that DelTriC correctly identifies anomalies located within clusters in the projected space. This advantage arises from its key feature of back-projection, which resolves anomalies in the original high-dimensional space.

\begin{table}[htbp]
\centering
\caption{Comparison of Clustering Algorithms for Anomaly Detection (F1 score)}
\small
\label{tab:synthetic_anomaly}
\begin{tabular}{lllll}
\toprule
$n_{\text{points}}$ & dim & DBSCAN & DelTriC & HDBSCAN \\
\midrule
500   & 5   & 0.16 & \textbf{0.62} & 0.33 \\
500   & 20  & 0.34 & \textbf{0.67} & 0.33 \\
500   & 50  & 0.14 & \textbf{0.73} & 0.20 \\
5000  & 5   & 0.20 & \textbf{0.42} & 0.16 \\
5000  & 20  & 0.28 & \textbf{0.68} & 0.29 \\
5000  & 50  & 0.30 & \textbf{0.82} & 0.35 \\
10000 & 5   & 0.18 & \textbf{0.46} & 0.16 \\
10000 & 20  & 0.31 & \textbf{0.73} & 0.34 \\
10000 & 50  & 0.28 & \textbf{0.77} & 0.30 \\
\bottomrule
\end{tabular}
\end{table}


In summary, DelTriC demonstrates clear advantages in the following scenarios:

\begin{itemize}
    \item High-dimensional spaces, where traditional algorithms such as HDBSCAN degrade in anomaly detection.
    \item Irregular cluster shapes (e.g., moons), where centroid-based methods like k-means fail.
\end{itemize}

DelTriC presents a compelling alternative to conventional clustering algorithms. Its hybrid approach – combining projection, triangulation, and back-mapping – enables both scalability and accuracy, making it especially suitable for complex, high-dimensional, and irregular data distributions.

\section{Performance on Real-World Data}

To evaluate the effectiveness of clustering algorithms on real-world datasets, we conducted experiments using five methods: \textbf{DelTriC} (our newly proposed algorithm), \textbf{HDBSCAN}, \textbf{KMeans}, \textbf{MiniBatchKMeans}, and \textbf{Spectral Clustering}. The datasets used include \textit{AG News}, \textit{DBPedia 14}, \textit{Emotion}, and \textit{CIFAR-10}, each with varying complexity and label distributions. The test split sizes for these datasets are present in table~\ref{tab:dataset_embedding_overview}

\begin{table*}[htbp]
\centering
\small 
\caption{Overview of Datasets and Embeddings Used. Embeddings: all-MiniLM-L6-v2~\cite{allminilm}, CLIP (ViT-B/32)~\cite{clip}.}
\label{tab:dataset_embedding_overview}
\begin{tabular}{l|c|l}
\toprule
Dataset & Samples & Embedding Used \\
\midrule
DBPedia-14~\cite{zhang2015character}  & 10,000       & all-MiniLM-L6-v2 \\
Emotion~\cite{emotion}             & 2,000        & all-MiniLM-L6-v2 \\
AG News~\cite{zhang2015character}             & 7,600        & all-MiniLM-L6-v2 \\
News20~\cite{news20}             & 5,500        & all-MiniLM-L6-v2 \\
CLINC OOS (plus)~\cite{clinc}   & 5,500        & all-MiniLM-L6-v2 \\
CIFAR-10~\cite{cifar10}            & 10,000       & CLIP (ViT-B/32) \\
PBMC~\cite{pbmc68k}    & 724          & None (numerical) \\
MNIST~\cite{mnist}                 & 10,000       & CLIP (ViT-B/32) \\
\bottomrule
\end{tabular}
\end{table*}

The clustering performance was measured using \textbf{Adjusted Rand Index (ARI)} and \textbf{Normalized Mutual Information (NMI)}. Table~\ref{tab:clustering_performance} summarizes the results.


\begin{table}[htbp]
\centering
\small
\caption{Clustering Performance on Real-World Datasets (ARI \textbar NMI).}
\label{tab:clustering_performance}
\begin{tabular}{llll} 
\toprule
Dataset & DBSCAN & DelTriC & HDBSCAN \\
\midrule
AG News        & 0.55 \textbar 0.56 & 0.51 \textbar 0.52 & 0.31 \textbar 0.49 \\
CIFAR-10       & 0.72 \textbar 0.76 & \textbf{0.73} \textbar \textbf{0.76} & 0.71 \textbar 0.75 \\
CLINC          & 0.58 \textbar 0.88 & 0.53 \textbar 0.87 & 0.60 \textbar 0.88 \\
DBPedia 14     & 0.64 \textbar 0.75 & 0.64 \textbar 0.75 & 0.58 \textbar 0.73 \\
Emotion        & 0.03 \textbar 0.07 & 0.05 \textbar 0.07 & 0.03 \textbar \textbf{0.08} \\
MNIST          & \textbf{0.85} \textbar \textbf{0.84} & 0.83 \textbar 0.83 & 0.78 \textbar 0.80 \\
PBMC           & 0.52 \textbar 0.60 & 0.49 \textbar 0.56 & 0.51 \textbar \textbf{0.63} \\
PBMC Lou.   & \textbf{0.61} \textbar 0.75 & 0.58 \textbar 0.73 & 0.58 \textbar 0.74 \\
\midrule
Dataset & KMeans & Spectral & \\ 
\midrule
AG News        & \textbf{0.63} \textbar \textbf{0.59} & 0.26 \textbar 0.35 & \\
CIFAR-10       & 0.63 \textbar 0.71 & 0.59 \textbar 0.74 & \\
CLINC          & \textbf{0.67} \textbar \textbf{0.90} & 0.07 \textbar 0.72 & \\
DBPedia-14     & \textbf{0.74} \textbar \textbf{0.81} & 0.58 \textbar 0.76 & \\
Emotion        & \textbf{0.05} \textbar 0.06 & 0.01 \textbar 0.03 & \\
MNIST          & 0.36 \textbar 0.47 & 0.61 \textbar 0.77 & \\
PBMC           & \textbf{0.58} \textbar 0.64 & 0.37 \textbar 0.59 & \\
PBMC Lou.   & 0.55 \textbar 0.75 & 0.58 \textbar \textbf{0.75} & \\
\bottomrule
\end{tabular}
\end{table}

The results in Table~\ref{tab:clustering_performance} indicate that KMeans achieves strong performance on several datasets (e.g., AG News, DBPedia). This suggests that, in such cases, the underlying data distribution resembles relatively well-separated, “blob-like” clusters in the embedding space. In these scenarios, more advanced density-based or graph-based algorithms mainly offer only the practical advantage of not requiring the number of clusters to be specified in advance.

In contrast, for datasets with more complex cluster structures, such as MNIST or CIFAR-10, the newer methods exhibit their strengths. DelTriC performs comparably to DBSCAN and HDBSCAN; however, this comparison is only approximate, since the evaluation must also account for the number of anomalies detected by each algorithm. A lower number of detected anomalies typically leads to higher ARI or NMI scores, as anomalies are not labeled in these benchmark datasets. Conversely, when an algorithm identifies many anomalies, there is no clear evidence that these truly correspond to outliers. The anomaly counts for each method are reported in Table~\ref{tab:anomaly_detection}.

\begin{table}[htbp]
\centering
\small
\caption{Number of anomalies detected on real-world datasets.}
\label{tab:anomaly_detection}
\begin{tabular}{l|c|c|c}
\toprule
Dataset & DBSCAN & DelTriC & HDBSCAN \\
\midrule
AG News & 154 & 215 & 41 \\
CIFAR-10 & 787 & 359 & 969 \\
CLINC & 204 & 325 & 218 \\
DBPedia-14 & 582 & 658 & 771 \\
Emotion & 485 & 263 & 519 \\
MNIST & 286 & 291 & 922 \\
PBMC & 40 & 97 & 26 \\
PBMC Louvain & 38 & 41 & 32 \\
\bottomrule
\end{tabular}
\end{table}

Table~\ref{tab:anomaly_detection} also sheds light on why HDBSCAN often performs worse than DBSCAN and DelTriC, as it typically identifies a large proportion of the data as anomalies. A notable case for DelTriC is the DBPedia dataset, where it produces clusters of comparable quality to DBSCAN but detects a greater number of anomalies. It is worth noting that the ARI and NMI scores of DelTriC could potentially be improved by lowering the anomaly-sensitivity hyperparameter, since this would merge a larger fraction of anomalies into existing clusters and thereby yield higher scores. However, in these experiments we deliberately set this hyperparameter to its maximum value.

\section{Anomaly Detection}
In this section, we evaluate the robustness of clustering algorithms on anomaly-enriched variants of the MNIST and News20 datasets. For MNIST, we introduced 500 anomalies into the standard test set of 10,000 samples by applying a set of transformations to existing images. The transformations considered were Gaussian blur, occlusion, elastic deformation, salt-and-pepper noise, and rotation. The CLIP embeddings were used as in the prior MNIST tests. For the News20 dataset, anomalies were derived directly from the data: specifically, news titles that were labeled to a given category but simultaneously appeared as outliers with respect to the corresponding k-means cluster.

We report clustering performance in terms of Adjusted Rand Index (ARI) and Normalized Mutual Information (NMI), and anomaly detection performance in terms of precision, recall, and F1 score.


\begin{table}[htbp]
\centering
\caption{Clustering Performance on Anomaly-Enriched MNIST and News20 (ARI \textbar NMI). MNIST transformations: Gaussian Blur, Flip, Mask, Noise, Rotation. News20: Outliers.}
\small
\label{tab:clustering_anomaly}
\begin{tabular}{llll} 
\toprule
Dataset & DBSCAN & DelTriC & HDBSCAN \\
\midrule
Gauss. Blur & 0.81 \textbar \textbf{0.86} & 0.81 \textbar 0.83 & 0.81 \textbar 0.86 \\
Flip          & 0.63 \textbar 0.74 & \textbf{0.78} \textbar 0.78 & 0.73 \textbar \textbf{0.78} \\
Mask          & 0.67 \textbar 0.76 & \textbf{0.77} \textbar \textbf{0.77} & 0.75 \textbar 0.79 \\
Noise         & 0.81 \textbar \textbf{0.86} & \textbf{0.85} \textbar 0.84 & 0.81 \textbar 0.86 \\
Rotation      & 0.56 \textbar 0.72 & \textbf{0.75} \textbar 0.76 & 0.68 \textbar \textbf{0.76} \\
Outliers     & 0.45 \textbar 0.51 & 0.45 \textbar 0.52 & 0.42 \textbar 0.50 \\
\bottomrule
\end{tabular}
\end{table}

The results of the experiments are summarized in Table~\ref{tab:clustering_anomaly}. Interestingly, DBSCAN and HDBSCAN appear to be highly sensitive to anomaly enrichment, as their performance often drops considerably compared to the baseline on the original MNIST dataset ($ARI \sim 0.8$). In contrast, DelTriC exhibits greater stability under these perturbations.

To evaluate the anomaly detection capability of the clustering algorithms, we compare DelTriC and HDBSCAN using precision, recall, and F1 score across the different anomaly types, as reported in Table~\ref{tab:anomaly_detection_prec}. The overall scores remain low, likely due to the CLIP embeddings, which are often invariant to the applied transformations. Moreover, even prior to enrichment, the MNIST dataset already contained natural anomalies that are not explicitly labeled as such. Despite the generally low scores, DelTriC consistently outperforms HDBSCAN, indicating its ability to detect anomalous patterns while maintaining high clustering quality, as reflected in Table~\ref{tab:clustering_anomaly}.


\begin{table}[htbp]
\centering
\small
\caption{Anomaly Detection Performance (Precision \textbar Recall \textbar F1)}
\label{tab:anomaly_detection_prec}
\begin{tabular}{l l c}
\toprule
Anomaly Type & Method & Precision \textbar Recall \textbar F1 \\
\midrule
\multicolumn{3}{l}{\textbf{MNIST}} \\
Gaussian Blur          & DelTriC & \textbf{0.04 \textbar 0.03 \textbar 0.03} \\
                       & HDBSCAN & 0.00 \textbar 0.00 \textbar 0.00 \\
                       & DBSCAN  & 0.00 \textbar 0.00 \textbar 0.00 \\
Flip                   & DelTriC & 0.22 \textbar \textbf{0.18} \textbar \textbf{0.20} \\
                       & HDBSCAN & 0.18 \textbar 0.13 \textbar 0.15 \\
                       & DBSCAN  & \textbf{0.25} \textbar 0.12 \textbar 0.16 \\
Mask                   & DelTriC & \textbf{0.17 \textbar 0.14 \textbar 0.15} \\
                       & HDBSCAN & 0.12 \textbar 0.05 \textbar 0.07 \\
                       & DBSCAN  & 0.13 \textbar 0.05 \textbar 0.07 \\
Salt-Pepper Noise      & DelTriC & \textbf{0.14 \textbar 0.11 \textbar 0.12} \\
                       & HDBSCAN & 0.00 \textbar 0.00 \textbar 0.00 \\
                       & DBSCAN  & 0.00 \textbar 0.00 \textbar 0.00 \\
Rotation               & DelTriC & 0.10 \textbar \textbf{0.08} \textbar 0.09 \\
                       & HDBSCAN & 0.15 \textbar 0.06 \textbar 0.09 \\
                       & DBSCAN  & \textbf{0.22} \textbar 0.08 \textbar \textbf{0.12} \\
\midrule
\multicolumn{3}{l}{\textbf{News20}} \\
Outliers               & DelTriC & \textbf{0.57} \textbar 0.17 \textbar 0.27 \\
                       & HDBSCAN & 0.45 \textbar 0.23 \textbar 0.31 \\
                       & DBSCAN  & 0.48 \textbar \textbf{0.24} \textbar \textbf{0.32} \\
\bottomrule
\end{tabular}
\end{table}

\section{Limitations}
\label{sec:limitations}

While DelTriC shows promising performance across a variety of datasets, it also has several limitations. First, in some benchmark tests it underperforms compared to HDBSCAN and DBSCAN, particularly on the \textit{pbmc} dataset and on low-dimensional synthetic datasets (e.g., 5000 points in 5 dimensions). Second, as expected, it performs no better than HDBSCAN or DBSCAN on datasets that are inherently well-suited to centroid-based methods such as $k$-means. Finally, similar to other clustering algorithms, DelTriC struggles when manifolds lie closer to each other than the typical intra-manifold point distances. In such cases, the back-projection mechanism may degrade, as illustrated in our small demo experiment in Appendix~\ref{demo_degradation}. Another limitation arises from the use of 2-dimensional UMAP, which is discussed in more detail at the end of Appendix~\ref{app:deltric}.

\section{Conclusion}

DelTriC shows promising results in the clustering domain, particularly when anomalies must be detected alongside clusters. Although datasets suitable for benchmarking clustering combined with outlier detection are scarce, we performed several tests on both synthetic and real-world data, demonstrating DelTriC’s exceptional performance. Moreover, the back-projection technique developed for DelTriC may have applications beyond the clustering domain.

\section{Acknowledgement}
This research was partially supported by DisAi, a project funded by the European Union under the Horizon Europe, \href{https://doi.org/10.3030/101079164}{GA No. 101079164} and was partially funded by European Union, under the project lorAI, \href{https://doi.org/10.3030/101136646}{GA No. 101136646}.

\bibliography{bibliography}

\appendix

\section{Theoretical Motivation for DelTriC: High-Dimensional Preservation via Back-Projection}
\label{theory}

Dimensionality reduction techniques such as principal component analysis (PCA) are widely used for visualizing and preprocessing high-dimensional data. However, it is well established that these methods can significantly distort important topological and geometric properties, including pairwise distances, local densities, and neighborhood structures~\cite{biorxiv2025}. Non-linear approaches such as UMAP address some of these limitations by focusing on the preservation of local relationships, yet they too are subject to information loss and may fail to fully maintain the global structure of the original data~\cite{chang2025survey}. In this section, we provide a theoretical motivation for the DelTriC algorithm, emphasizing how its back-projection step helps recover high-dimensional structure that is typically lost during projection.

\subsection{Projection-Induced Distortion of Cluster Structure}

Let $X \subset \mathbb{R}^d$ denote the original high-dimensional data, and let $\pi: \mathbb{R}^d \rightarrow \mathbb{R}^2$ be a linear projection, such as PCA. For two clusters $C_1, C_2 \subset X$, it is possible that:
\[
|| x - y || \gg ||\pi(x) - \pi(y)|| \quad \text{for } x \in C_1, y \in C_2,
\]
indicating that well-separated clusters in the original space may appear artificially merged in the projection. This loss of injectivity can cause false positives during clustering in the low-dimensional space. DelTriC avoids this issue by using the projected space only to construct a triangulation skeleton, which is then evaluated in the original high-dimensional space.

\subsection{Delaunay Triangulation Preserves Local Topology}

In $\mathbb{R}^2$, Delaunay triangulation captures the local adjacency structure of points more effectively than $k$-nearest neighbors, especially in irregular distributions. Let $\mathcal{T}$ denote the triangulation of $\pi(X)$, and let each triangle $t = \{x_i, x_j, x_k\}$ have an associated perimeter or circumradius $R(t)$. DelTriC removes large triangles under the assumption that such triangles often span low-density regions or boundaries between clusters.

Importantly, DelTriC evaluates $R(t)$ not in $\mathbb{R}^2$ but in the original $\mathbb{R}^d$:
\[
R_{\mathbb{R}^d}(t) = \max_{i,j \in t} ||x_i - x_j||_{\mathbb{R}^d}.
\]
This correction prevents triangles that appear small in the projection (due to PCA compression) from misleading the clustering algorithm.

\subsection{Manifold Hypothesis and Geodesic Distortion}

High-dimensional data often lies on a lower-dimensional manifold $\mathcal{M} \subset \mathbb{R}^d$. Let $d_{\mathcal{M}}(x, y)$ be the geodesic distance along $\mathcal{M}$. In many cases,
\[
d_{\mathcal{M}}(x, y) \gg ||x - y||_2,
\]
suggesting that two points close in Euclidean space may be distant on the manifold. Linear projections can exacerbate this by bringing distant points even closer, leading to incorrect cluster merging.

DelTriC helps mitigate this by back-projecting the triangle structure and re-evaluating neighborhood relationships using high-dimensional distances, providing a rough approximation of geodesic separability.

\subsection{Advantages Over Clustering in Projected Space}

Traditional clustering algorithms applied in the projected space suffer from three primary issues:
\begin{itemize}
    \item Loss of information about volume elements, degrading density estimation.
    \item Compression of inter-cluster margins, complicating boundary detection.
    \item Distorted neighborhood graphs, affecting hierarchical or spectral clustering.
\end{itemize}

DelTriC addresses these by:
\begin{enumerate}
    \item Efficiently estimating adjacency using 2D Delaunay triangulation.
    \item Validating cluster boundaries using high-dimensional triangle metrics.
    \item Combining the scalability of 2D projection with the fidelity of high-dimensional geometry.
\end{enumerate}

This hybrid structure allows DelTriC to outperform traditional projection-based clustering methods, especially in high-dimensional spaces with complex cluster shapes.

\section{DelTriC in a Nutshell}
\label{app:deltric}

For readers who prefer a concise, high-level textual description, we summarize the \textbf{DelTriC} algorithm as follows:
\begin{enumerate}
\item \textbf{Dimensionality Reduction:} Project the data into two dimensions using techniques such as UMAP or PCA.
\item \textbf{Triangulation:} Apply Delaunay triangulation to the 2-dim projection.
\item \textbf{Back-Projection:} Map the resulting triangles back into the original high-dimensional space.
\item \textbf{Cluster Formation:} Iteratively remove the largest triangles to form clusters.
\item \textbf{Cluster Merging:} Optionally, merge clusters by repeating the process on their centroids.
\item \textbf{Anomaly Merging:} Optionally, merge anomalies based on the triangulation obtained in step 2.
\end{enumerate}

In contrast to triangulation-based approaches discussed in Section~\ref{relatedwork}, which operate exclusively in the projected space, DelTriC reprojects the triangulated structure back into the original feature space. This back-projection enables more accurate clustering in high-dimensional settings. Cluster formation is driven by the removal of triangle edges according to their circumference.

The algorithm is governed by three key parameters:
\begin{itemize}
\item \texttt{prune$\_$param} controls the degree of triangle suppression during clustering, expressed as a z-score relative to the distribution of triangle sizes.
\item \texttt{merge$\_$param} determines the extent of cluster merging when centroids are sufficiently close, also expressed in z-scores.
\item \texttt{anomaly$\_$sensitivity} specifies how strictly anomalies are kept separate from clusters (with $1 =$ never merged, $0 =$ always merged).
\end{itemize}

\begin{figure}[htbp]
\centering
\begin{subfigure}[t]{0.48\textwidth}
    \centering
    \includegraphics[width=\linewidth]{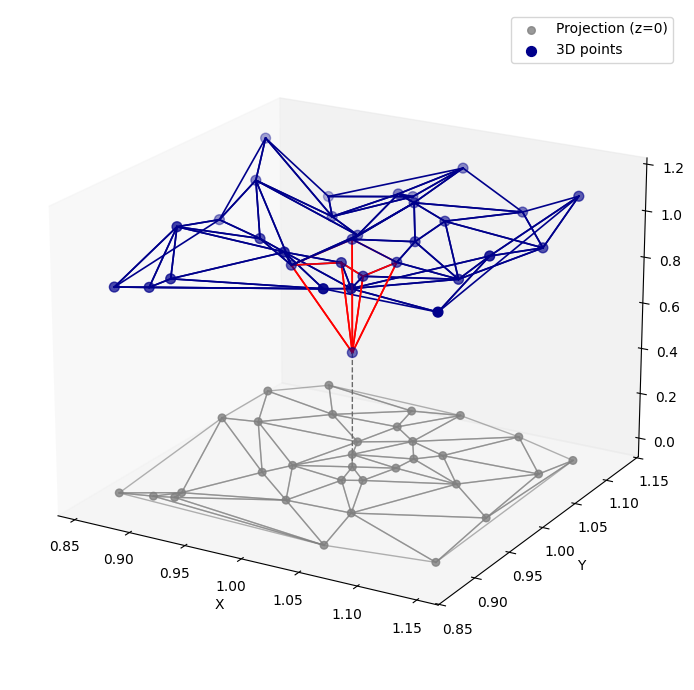}
    \caption{DelTriC method illustrated on 30 synthetic points. 
    The projection is shown in gray. 
    29 points roughly follow a manifold \textit{z=1}, with a single anomaly located at a lower position. 
    Although this anomaly appears embedded within the cluster in the projection plane, 
    DelTriC correctly identifies it through back-projection.}
    \label{fig:illustrative1}
\end{subfigure}
\hfill
\begin{subfigure}[t]{0.48\textwidth}
    \centering
    \includegraphics[width=\linewidth]{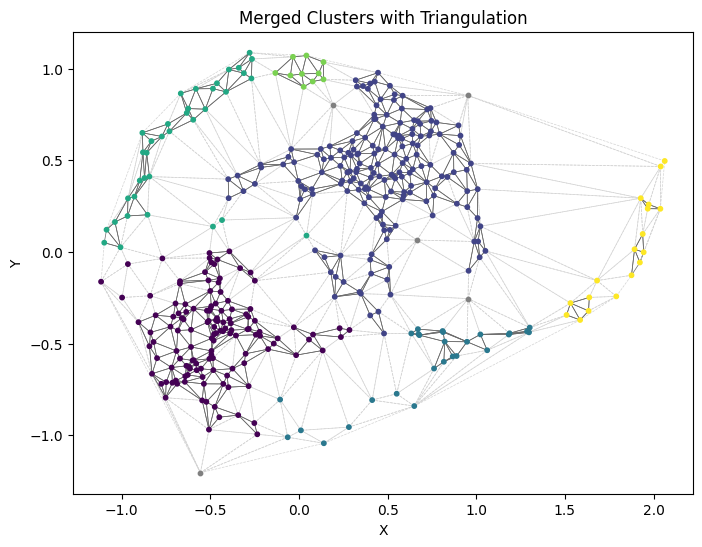}
    \caption{Clustering result on irregular shapes with the triangulation highlighted. 
    The pruned edges are depicted with dashed lines.}
    \label{fig:illustrative2}
\end{subfigure}
\caption{Illustrative examples: (a) anomaly detection with DelTriC, (b) clustering with triangulation.}
\label{fig:combined}
\end{figure}

To provide intuition for the back-projection step, we depict an illustrative scenario in Figure~\ref{fig:illustrative1}. The process of edge pruning is demonstrated in Figure~\ref{fig:illustrative2}.

\begin{figure}[htbp]
\centering
\includegraphics[width=\columnwidth]{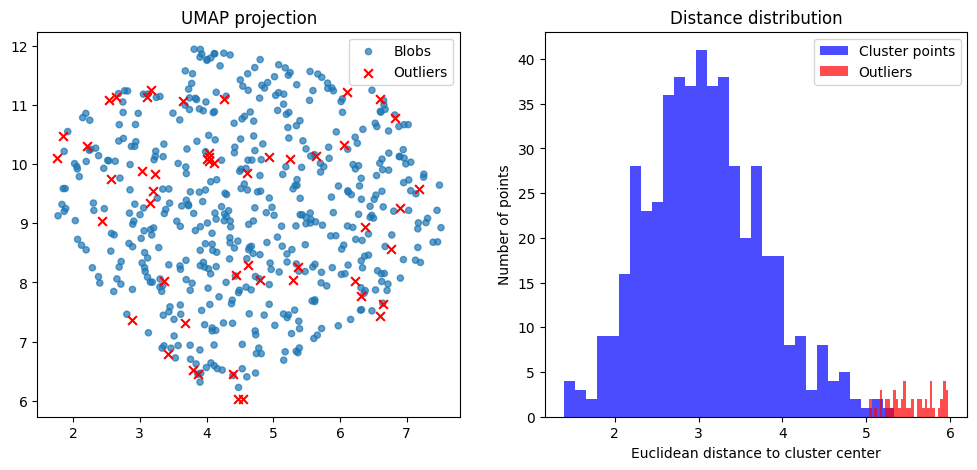}
\caption{A simple test to demonstrate that outliers can end up as internal points of a cluster after dimensionality reduction.}
\label{fig:demo_umap}
\end{figure}

Finally, a simple test can be implemented to highlight the necessity of back-projection, as shown in Figure~\ref{fig:demo_umap}, which illustrates the key innovation of DelTriC. The test generates a small Gaussian blob in 10-dimensional space and labels points on the tail of the distribution as anomalies (left panel). When projected into two dimensions with UMAP, these anomalies become embedded within the main cluster (right panel). This makes them undetectable for algorithms that rely solely on the projected space, whereas DelTriC successfully identifies them through back-projection.

\begin{figure}[htbp]
\centering
\includegraphics[width=\columnwidth]{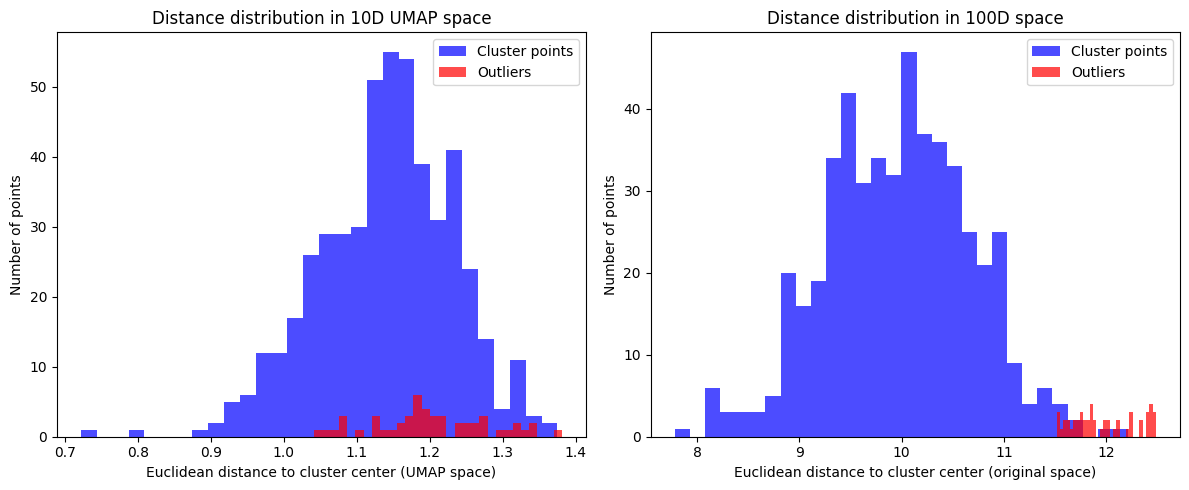}
\caption{A simple test to demonstrate that outliers can end up as internal points of a cluster after dimensionality reduction.}
\label{fig:demo_umap_2}
\end{figure}

The same issue persists in higher dimensions, as illustrated in Figure~\ref{fig:demo_umap_2}, where we reduce a 100-dimensional space to 10 dimensions. Since direct visualization of the 10-dimensional embedding is not feasible, the left plot was replaced with a histogram of Euclidean distances from the cluster center.

The choice of the target dimensionality for UMAP should be made carefully, depending on the expected intrinsic dimensionality of the manifolds present in the data. For DelTriC, we are constrained to use a 2-dimensional UMAP projection (with a possible extension to 3 dimensions), whereas for DBSCAN and HDBSCAN we employed a 10-dimensional projection to allow the methods to retain as much structural information from the high-dimensional space as possible. We acknowledge, however, that some high-dimensional datasets may in fact contain 2-dimensional manifolds. In such cases, a 10-dimensional UMAP projection may be less effective at reconstructing the underlying structure than a lower-dimensional embedding.

\section{DelTriC Pruning Degradation}
\label{demo_degradation}

\begin{figure}[htbp]
\centering
\includegraphics[width=\columnwidth]{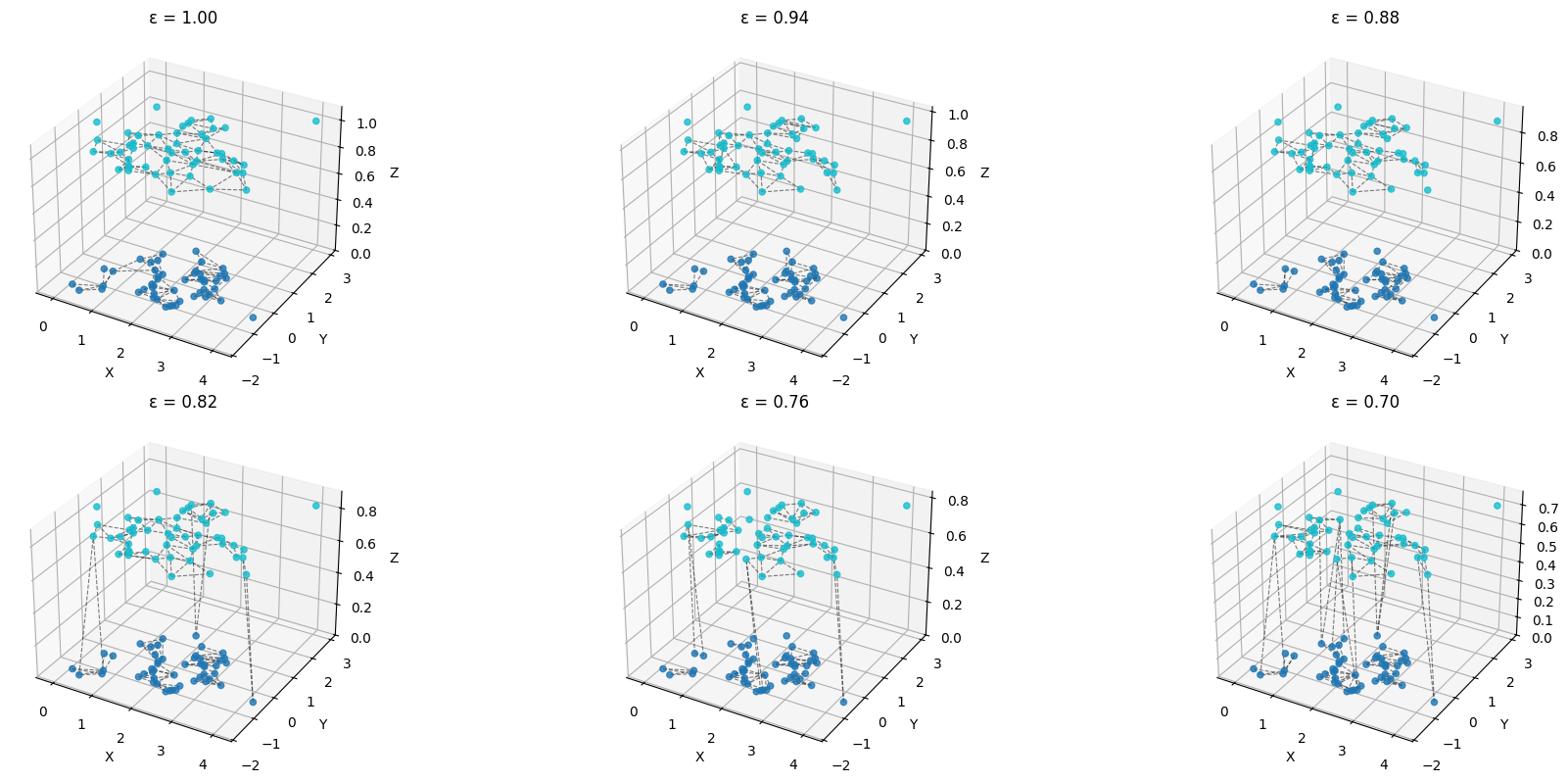}
\caption{A simple test to demonstrate pruning step degradation when manifolds get very close.}
\label{fig:demo}
\end{figure}

When manifolds are in close proximity, the pruning step of DelTriC may deteriorate. Figure~\ref{fig:demo} demonstrates this effect by systematically reducing the inter-manifold distance parameter $\epsilon$ and observing when inter-manifold edges are no longer removed. The data were generated with an average intra-manifold edge length of $0.88$. Results indicate that DelTriC remains effective for $\epsilon \approx 1.0$, but the pruning step progressively fails at smaller $\epsilon$ values, retaining inter-manifold edges.

\section{Ablation}

To better understand the contribution of individual components in the DelTriC clustering algorithm, we conducted an ablation study across three anomaly-enriched datasets: CIFAR-10, DBpedia-14 and MNIST dataset in the same manner as in previous sections. We evaluated six variants of the model:

\begin{itemize}
    \item \textbf{DelTriC}: Full model with all components.
    \item \textbf{Merge Off}: Merging disabled.
    \item \textbf{Proj Off}: Back-projection disabled.
    \item \textbf{Merge + Proj Off}: Both merging and back-projection disabled.
    \item \textbf{DelTriC Anom. Merged}: Full model with all components but all anomalies merged into existing clusters.
    \item \textbf{Proj Off Anom. Merged}: Back-projection disabled. Anomalies merged into existing clusters.
\end{itemize}

As shown in Table~\ref{tab:ablation}, disabling the back-projection component (DelTriC\_proj\_off) resulted in a very slight improvement in clustering performance, measured by ARI and NMI, for the MNIST dataset. However, this difference is not statistically significant and likely reflects normal fluctuations. Overall, the performance on MNIST is largely consistent across all DelTriC variants, highlighting the strong influence of Delaunay triangulation and UMAP in this case. In contrast, for CIFAR-10, and particularly for DBpedia-14, all ablated variants of DelTriC exhibit markedly worse performance, demonstrating the necessity of retaining all DelTriC components. As expected, clustering performance further improves when all anomalies are merged into existing clusters, as discussed in previous sections. Moreover, the key strength of DelTriC lies in anomaly detection, and disabling the back-projection component substantially undermines this capability.



\begin{table}[htbp]
\centering
\caption{Transposed Clustering Performance of DelTriC Variants on New Datasets}
\label{tab:ablation}
\begin{tabular}{l|c|c|c}
\toprule
\textbf{Metric / Method} & \textbf{CIFAR-10} & \textbf{DBpedia-14} & \textbf{MNIST} \\
\midrule
ARI -- DelTriC & 0.73 & \textbf{0.65} & 0.83 \\
ARI -- Merge Off & 0.72 & 0.56 & 0.83 \\
ARI -- Proj Off & 0.70 & 0.54 & \textbf{0.85} \\
ARI -- Merge+Proj Off & 0.69 & 0.55 & 0.85 \\
ARI -- DelTriC Anom. Merged & \textbf{0.74} & 0.68 & 0.84 \\
ARI -- Proj Off Anom. Merged & 0.69 & 0.55 & 0.85 \\
\hline
NMI -- DelTriC & 0.76 & 0.75 & 0.83 \\
NMI -- Merge Off & 0.76 & 0.72 & 0.83 \\
NMI -- Proj Off & 0.75 & 0.71 & 0.83 \\
NMI -- Merge+Proj Off & 0.74 & 0.71 & \textbf{0.84} \\
NMI -- DelTriC Anom. Merged & \textbf{0.78} & \textbf{0.78} & 0.84 \\
NMI -- Proj Off Anom. Merged & 0.75 & 0.72 & 0.84 \\
\bottomrule
\end{tabular}
\end{table}

\section{Observations}

\begin{itemize}
    \item \textbf{DelTriC}, despite using only \textbf{2 UMAP dimensions} due to its triangulation-based nature, demonstrated competitive performance across datasets. Notably, on \textit{CIFAR-10}, it revealed internal structure that was visually evident in the UMAP projection, suggesting effective back-projection capabilities.
    \item All other algorithms used \textbf{10-dimensional UMAP embeddings}, which may have contributed to their performance, especially on high-dimensional datasets.
    \item \textbf{KMeans} and \textbf{Spectral Clustering} benefited from having the number of clusters (\textit{k}) preset to match the number of ground-truth labels, giving them an advantage in alignment with true classes.
    \item \textbf{HDBSCAN}, being density-based and non-parametric, showed moderate performance but struggled on datasets with less distinct cluster boundaries.

\end{itemize}

\section{Reproducibility}






Additionally to the information given in the main text we state that:

\begin{itemize}
    \item All experiments were run with random seed 42 for all libraries and frameworks.
    \item We use the official train/test/validation splits as provided by HuggingFace for all datasets with no
    additional filtering.
    \item HDBSCAN, DBSCAN and DelTriC got exactly 100 probes in Optuna no matter of how many hyper-parameters used in order to guarantee fairness across the three algorithms.
\end{itemize}

We also list all hyper-parameters for each for each experiment found by Optuna in three separate tables: HDBSCAN in Table~\ref{tab:params_dbscan}, DBSCAN in Table~\ref{tab:params_hdbscan} and DelTriC in Table~\ref{tab:params_deltric}.

In order to easily reproduce the reported benchmarks and use the DelTriC, it was made public with PyPI and one can easily import it into a Python environment. The following snippet demonstrates a minimal setup; however, users are encouraged to optimize the hyper-parameters for their specific use cases.

\begin{verbatim}
% pip install deltric==0.1.0
\end{verbatim}

\begin{verbatim}
from sklearn.datasets import make_blobs
from deltric.deltric import DelTriC

# Generate synthetic data
X, true_labels = make_blobs(
    n_samples=500, centers=5, n_features=10,
    cluster_std=5, random_state=42
)

# Run DelTriC with default hyperparameters
dt = DelTriC(
    prune_param=0.3,
    merge_param=-0.8,
    dim_reduction='umap',
    back_proj=True,
    anomaly_sensitivity=0.99
)
labels = dt.fit_predict(X)
\end{verbatim}

For real-world experiments, datasets can be downloaded directly from HuggingFace. For instance, a dataset designed specifically for anomaly detection can be accessed as follows:

\begin{verbatim}
from datasets import load_dataset

# Login using, e.g., `huggingface-cli login` if required
ds = load_dataset("TomasJavurek/mnist-augmented")
\end{verbatim}

Finally we list all the versions of all the 3rd party dependencies in Table~\ref{tab:dependencies}

\begin{table}[htbp]
\centering
\begin{tabular}{ll}
\hline
\textbf{Package} & \textbf{Version} \\
\hline
Pillow & 11.0.0 \\
datasets & 4.0.0 \\
hdbscan & 0.8.40 \\
matplotlib & 3.10.5 \\
networkx & 3.3 \\
numpy & 2.1.2 \\
optuna & 4.5.0 \\
pandas & 2.3.1 \\
scanpy & 1.11.4 \\
scipy & 1.16.1 \\
seaborn & 0.13.2 \\
sentence\_transformers & 5.1.0 \\
scikit-learn & 1.7.1 \\
statsmodels & 0.14.5 \\
torch & 2.5.1+cu121 \\
torchvision & 0.20.1+cu121 \\
transformers & 4.55.2 \\
umap-learn & 0.5.9.post2 \\
\hline
\end{tabular}
\caption{List of Python package dependencies and their versions.}
\label{tab:dependencies}
\end{table}

\begin{table}[htbp]
\centering
\begin{tabular}{lcc}
\hline
benchmark & cluster\_selection\_epsilon & cluster\_selection\_method \\
\hline
10000x20 & 0.325499 & leaf \\
10000x5 & 0.313546 & leaf \\
10000x50 & 0.046700 & eom \\
5000x20 & 0.358032 & leaf \\
5000x5 & 0.346119 & leaf \\
5000x50 & 0.106094 & eom \\
500x20 & 0.456874 & leaf \\
500x5 & 0.489206 & leaf \\
500x50 & 0.405111 & eom \\
ag\_news & 0.546954 & leaf \\
cifar10 & 0.229134 & leaf \\
clinc\_oos & 0.169253 & leaf \\
dbpedia\_14 & 0.239174 & leaf \\
emotion & 0.483876 & leaf \\
mnist & 0.215158 & leaf \\
mnist-blur & 0.207781 & leaf \\
mnist-flip & 0.258924 & leaf \\
mnist-mask & 0.312976 & eom \\
mnist-noise & 0.298458 & leaf \\
mnist-rotation & 0.229547 & leaf \\
news\_20 & 0.385409 & leaf \\
news\_20\_anom & 0.361301 & leaf \\
pbmc68k\_cell\_type & 0.350096 & eom \\
pbmc68k\_louvain & 0.251190 & eom \\
\hline
\end{tabular}
\caption{The fine-tuned hyper-parameters for HDBSCAN.}
\label{tab:params_hdbscan}
\end{table}

\begin{table}[htbp]
\centering
\begin{tabular}{lcc}
\hline
benchmark & eps & min\_samples \\
\hline
10000x20 & 0.380911 & 18.0 \\
10000x5 & 0.401392 & 17.0 \\
10000x50 & 0.336275 & 15.0 \\
5000x20 & 0.400498 & 14.0 \\
5000x5 & 0.508457 & 20.0 \\
5000x50 & 0.332802 & 10.0 \\
500x20 & 0.551976 & 11.0 \\
500x5 & 0.712255 & 19.0 \\
500x50 & 0.558232 & 17.0 \\
ag\_news & 0.490599 & 18.0 \\
cifar10 & 0.260172 & 13.0 \\
clinc\_oos & 0.183208 & 6.0 \\
dbpedia\_14 & 0.293570 & 19.0 \\
emotion & 0.529342 & 12.0 \\
mnist & 0.310799 & 16.0 \\
mnist-blur & 0.313174 & 16.0 \\
mnist-flip & 0.376469 & 20.0 \\
mnist-mask & 0.315588 & 18.0 \\
mnist-noise & 0.354388 & 18.0 \\
mnist-rotation & 0.338288 & 16.0 \\
news\_20 & 0.407158 & 11.0 \\
news\_20\_anom & 0.439568 & 14.0 \\
pbmc68k\_cell\_type & 0.570197 & 20.0 \\
pbmc68k\_louvain & 0.580796 & 20.0 \\
\hline
\end{tabular}
\caption{The fine-tuned hyper-parameters for DBSCAN.}
\label{tab:params_dbscan}
\end{table}

\begin{table}[htbp]
\centering
\begin{tabular}{lcc}
\hline
benchmark & prune\_param & merge\_param \\
\hline
10000x20 & 0.871327 & -1.068806 \\
10000x5 & 0.715483 & -1.385047 \\
10000x50 & 0.840101 & -0.753863 \\
5000x20 & 0.982618 & -1.271216 \\
5000x5 & 0.592910 & -1.062261 \\
5000x50 & 0.775523 & -1.217083 \\
500x20 & 0.540334 & -1.404835 \\
500x5 & 0.507286 & -1.156454 \\
500x50 & 0.509480 & -1.667501 \\
ag\_news & 1.499065 & -0.188278 \\
cifar10 & 1.514234 & -1.176349 \\
clinc\_oos & 1.499668 & -0.942711 \\
dbpedia\_14 & 1.319012 & -0.464040 \\
emotion & 0.614144 & -0.477708 \\
mnist & 1.528077 & -1.428322 \\
mnist-blur & 0.142756 & -0.657291 \\
mnist-flip & 0.310280 & -0.714070 \\
mnist-mask & 0.314879 & -0.793945 \\
mnist-noise & 0.199587 & -0.952094 \\
mnist-rotation & 0.213082 & -0.647257 \\
news\_20 & 0.285699 & -0.689650 \\
news\_20\_anom & 0.721050 & -0.794214 \\
pbmc68k\_cell\_type & 0.664678 & -0.932849 \\
pbmc68k\_louvain & 1.256253 & -0.887629 \\
\hline
\end{tabular}
\caption{The fine-tuned hyper-parameters for DelTriC.}
\label{tab:params_deltric}
\end{table}

\section{Future Directions for DelTriC}

We identify several avenues for further development and exploration of DelTriC:  

\begin{itemize}
    \item \textbf{Reduction of hyperparameters.} The parameters $merge\_param$ and $prune\_param$ appear to be correlated and could potentially be consolidated into a single parameter. Moreover, an optimal value might be inferred directly from the input dataset prior to DelTriC execution, providing an additional practical advantage.
    \item \textbf{Extension to higher dimensions.} While triangulation is currently performed in two-dimensional space, the algorithm could be extended to three dimensions, potentially enhancing its sensitivity to anomalies.
    \item \textbf{Tuning of statistical components.} Certain statistical operations within the algorithm could be refined; for example, using the mode or median instead of the mean might yield better results in specific scenarios.
    \item \textbf{Alternative applications.} DelTriC could potentially be adapted for anomaly detection, serving as an alternative to clustering-based methods.
\end{itemize}

\subsection{Summary}

In summary, DelTriC outperforms previous triangulation-based methods by combining advanced density estimation, density-based partitioning, robust noise handling, and – crucially – an explicit back-projection to the original multidimensional space. These innovations enable DelTriC to handle adjacent clusters, heterogeneous densities, and complex spatial domains with greater accuracy and interpretability than prior approaches.

\section{Time Complexity}
\label{time_complexity}

\begin{figure}[htbp]
    \centering
    \begin{subfigure}[b]{0.48\textwidth}
        \centering
        \includegraphics[width=\textwidth]{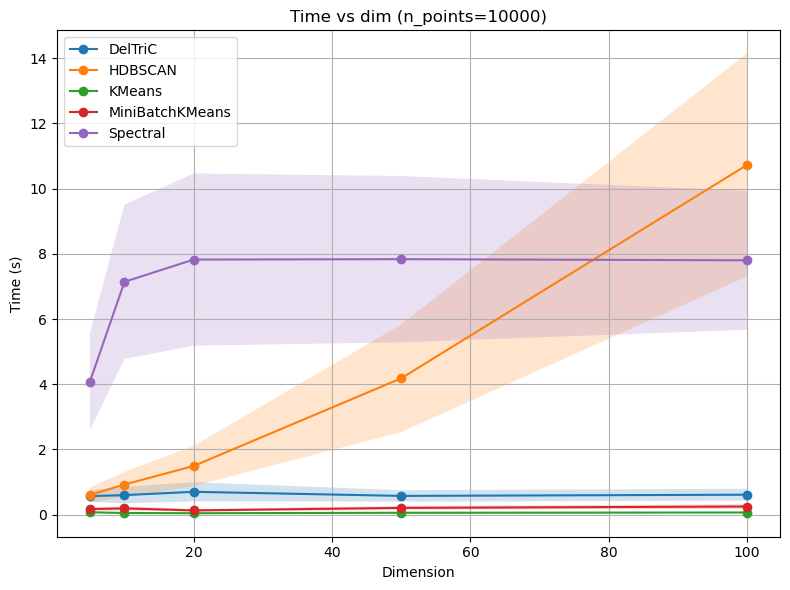}
        \subcaption{Computation time for clustering algorithms with fixed number of points and varying dimensionality.}
    \end{subfigure}
    \hfill
    \begin{subfigure}[b]{0.48\textwidth}
        \centering
        \includegraphics[width=\textwidth]{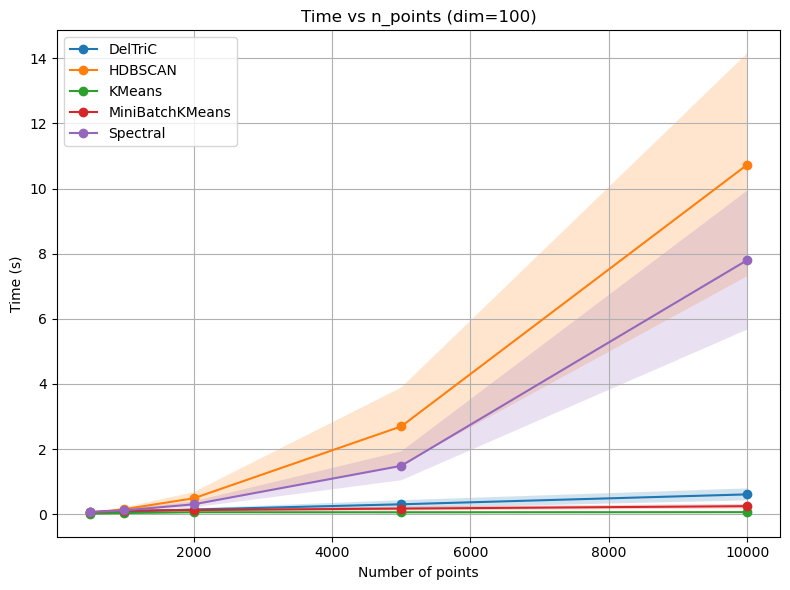}
        \subcaption{Computation time for clustering algorithms with fixed dimensionality and varying number of points.}
    \end{subfigure}
    \caption{Computation time of clustering algorithms on synthetic datasets without dimensionality reduction. DelTriC uses PCA for projection.}
    \label{fig:times}
\end{figure}

Figure~\ref{fig:times} illustrates the computational performance of various clustering algorithms on synthetic datasets, excluding UMAP from the pipeline. While UMAP typically dominates runtime due to its complexity, this experiment isolates the raw performance of clustering algorithms. For DelTriC, dimensionality reduction is an integral part of the algorithm; hence, PCA was used as a lightweight alternative to maintain fairness in comparison.

The results clearly show that modern clustering algorithms – when applied without dimensionality reduction – struggle with scalability in high-dimensional spaces and large datasets. In contrast, DelTriC maintains robust performance under these conditions, demonstrating its efficiency and suitability for high-dimensional clustering tasks.

\section{Example of Performance on Irregular Shapes}

\begin{figure}[htbp]
    \centering
    \includegraphics[width=0.9\textwidth]{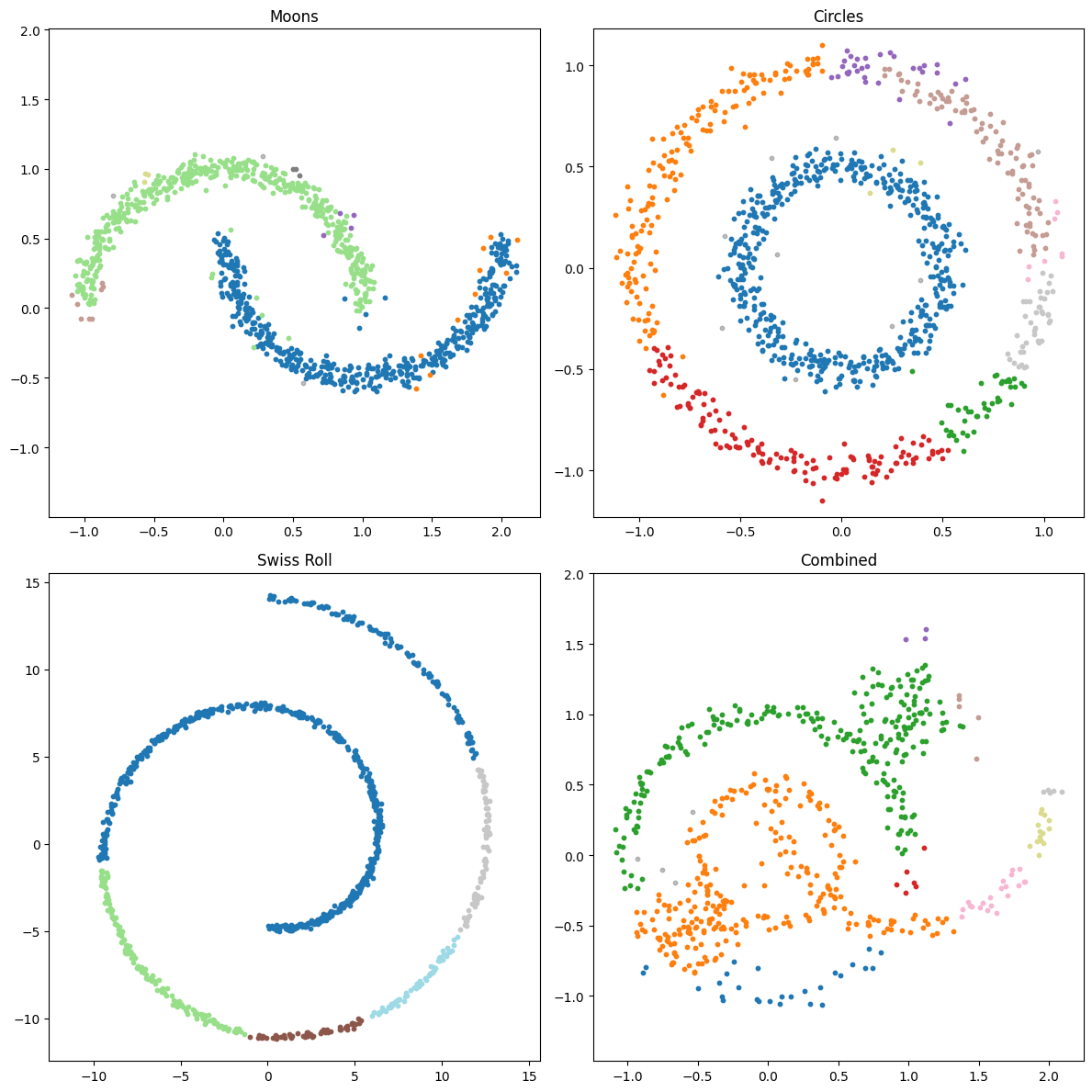}
    \caption{DelTriC performance on irregular cluster shapes.}
    \label{fig:irregular}
\end{figure}

Additionally, Figure~\ref{fig:irregular} showcases DelTriC's performance on irregular cluster shapes in two-dimensional space. In these cases, DelTriC performs comparably to HDBSCAN and significantly better than k-means. 

\section{Example of Clustering Result with Detected Outliers}

\begin{figure}[htbp]
\centering
\includegraphics[width=0.8\textwidth]{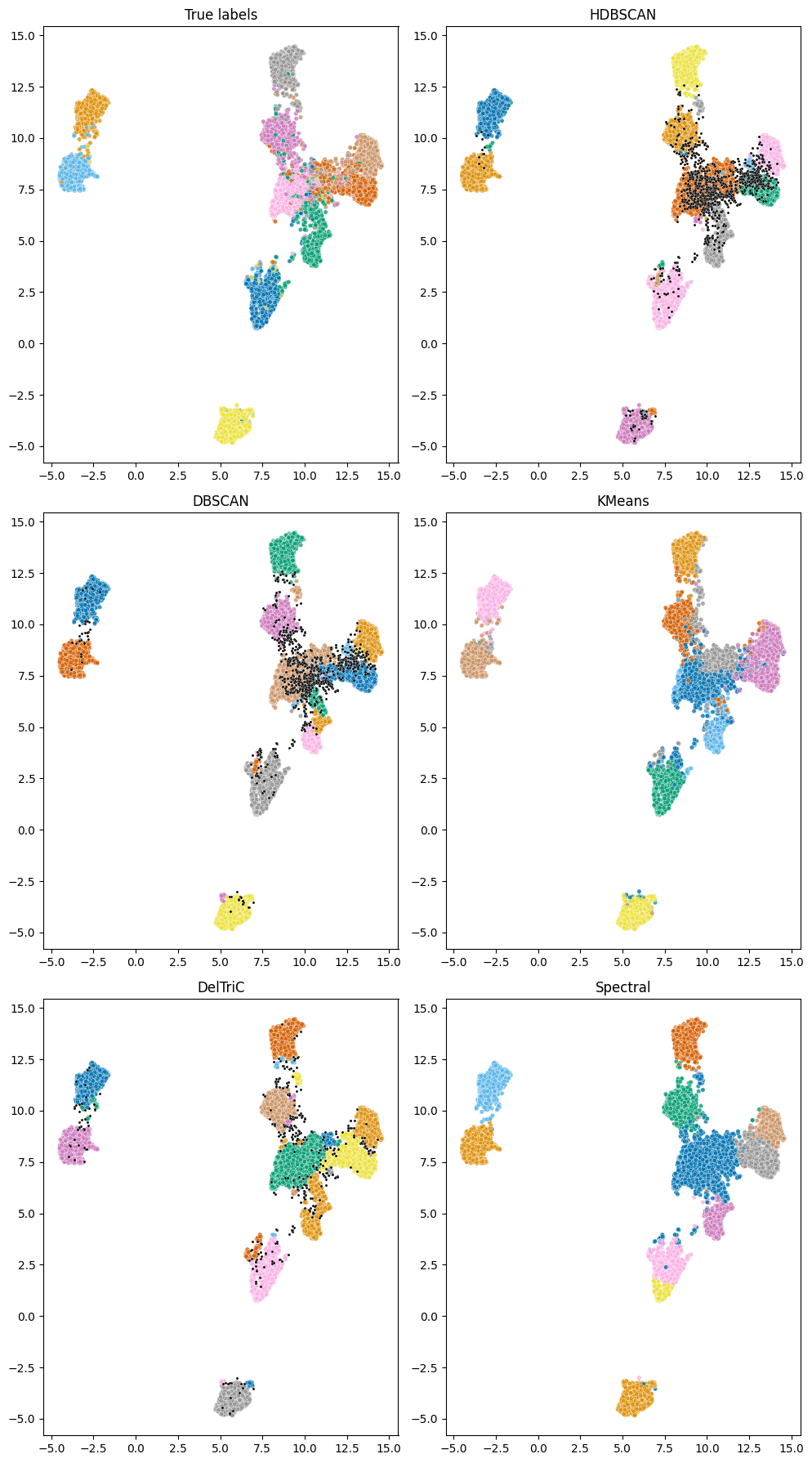}
\caption{UMAP projection of CIFAR-10 with clustering results from all algorithms.}
\label{fig:umap_cifar10}
\end{figure}

Figure~\ref{fig:umap_cifar10} illustrates an example benchmark on the CIFAR-10 dataset, where DelTriC outperformed both DBSCAN and HDBSCAN. Since anomalies are not explicitly annotated in the CIFAR-10 dataset, we represented them as smaller circular markers in dark gray. The plot shows that DelTriC identified fewer anomalies compared to DBSCAN and HDBSCAN, which is consistent with the results reported in Table~\ref{tab:anomaly_detection}.

\end{document}